\documentclass{article}
\usepackage{}
\usepackage{ijcai20}

\usepackage{times}
\usepackage{soul}
\usepackage{url}
\usepackage[hidelinks]{hyperref}
\usepackage[utf8]{inputenc}
\usepackage[small]{caption}
\usepackage{graphicx}
\usepackage{subfigure}
\usepackage{amsmath}
\usepackage{amsthm}
\usepackage{amsfonts}       
\usepackage{booktabs}
\usepackage{multirow}
\usepackage{threeparttable}
\usepackage{arydshln}
\usepackage{color}
\title{WaveSNet: Wavelet Integrated Deep Networks for Image Segmentation}

\author{
Qiufu Li
\and
Linlin Shen
\affiliations
Computer Vision Institute, Shenzhen University, China
\emails
\{liqiufu, llshen\}@szu.edu.cn
}

\begin{document}

\maketitle

\begin{abstract}
  In deep networks, the lost data details significantly degrade the performances of image segmentation.
  In this paper, we propose to apply Discrete Wavelet Transform (DWT) to extract the data details during feature map down-sampling,
  and adopt Inverse DWT (IDWT) with the extracted details during the up-sampling to recover the details.
  We firstly transform DWT/IDWT as general network layers,
  which are applicable to 1D/2D/3D data and various wavelets like Haar, Cohen, and Daubechies, etc.
  Then, we design wavelet integrated deep networks for image segmentation (WaveSNets)
  based on various architectures, including U-Net, SegNet, and DeepLabv3+.
  Due to the effectiveness of the DWT/IDWT in processing data details,
  experimental results on CamVid, Pascal VOC, and Cityscapes show
  that our WaveSNets achieve better segmentation performances than their vanilla versions.

\end{abstract}

\section{Introduction}
\begin{figure}[t]
	\centering
	\includegraphics*[scale=0.625, viewport=73 83 454 469]{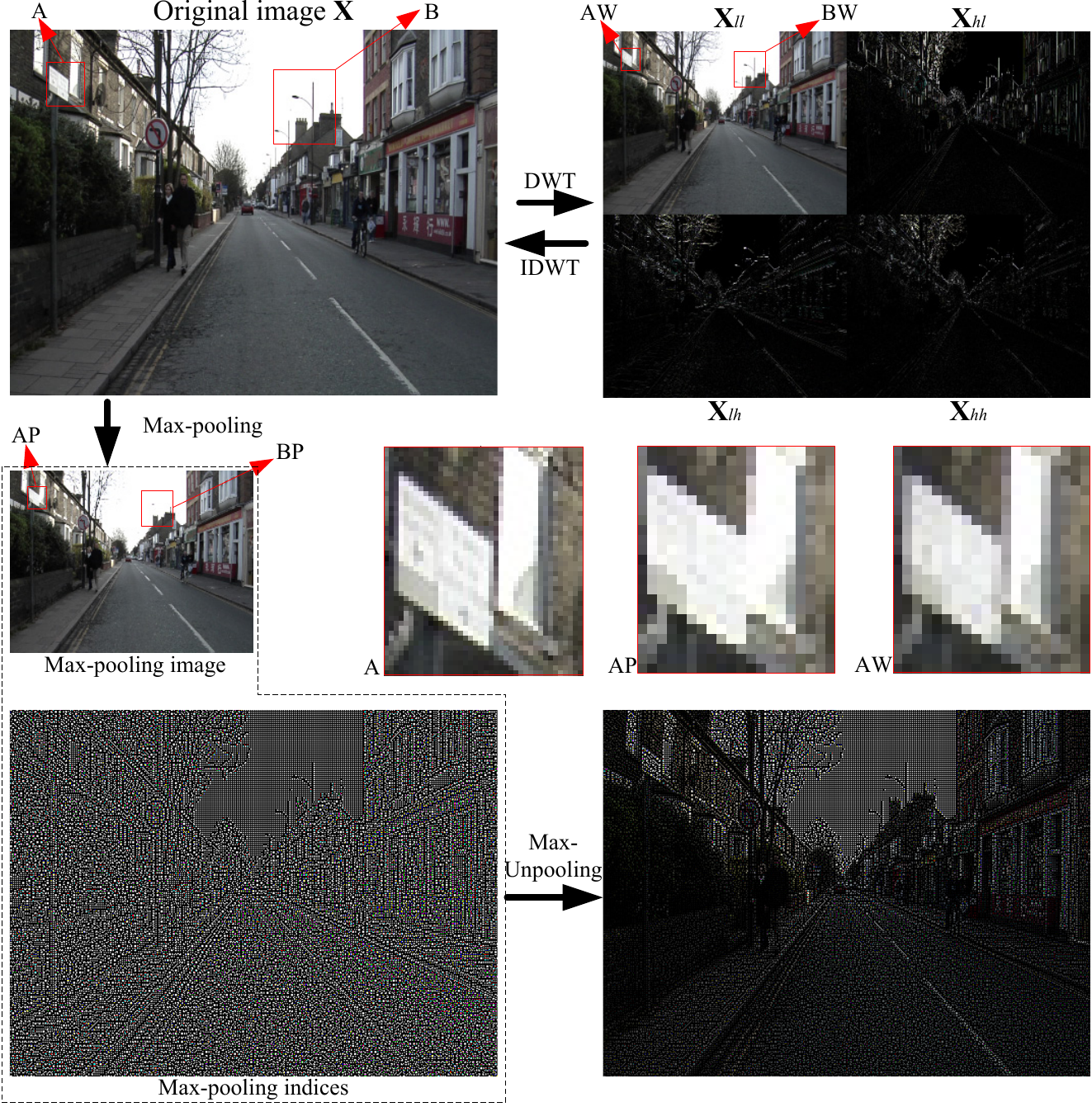}
	\caption{Comparison of max-pooling, max-unpooling, and wavelet transforms.
	Max-pooling is a common down-sampling operation in the deep networks,
	which easily removes the data details (the window boundary in area A) and breaks the object structure (the pole in area B).
	Based on the pooling indices, max-unpooling recovers the data resolution, but not the details.
	DWT keeps the object structure and saves the data details,
	while IDWT precisely recovers the original data using the DWT output.
	}\label{fig_MaxPooling_DWT_IDWT}
\end{figure}
Due to the advantage of deep network in extracting high-level features,
deep learning has achieved high performances in various tasks, particular in the computer vision.
However, the current deep networks are not good at extracting and processing data details.
While deep networks with more layers are able to fit more functions,
the deeper networks are not always associated with better performances \cite{he2016deep_resnet}.
An important reason is that the details will be lost as the data flow through the layers.
In particular, the lost data details significantly degrade the performacnes of the deep networks for the image segmentation.
Various techniques, such as condition random field,
\`atrous convolution \cite{chen2014semantic_deeplabv1,chen2018encoder_deeplabv3+}, PointRend \cite{kirillov2019pointrend},
are introduced into the deep networks to improve the segmentation performance.
However, these techniques do not explicitly process the data details.

Wavelets \cite{daubechies1992ten,mallat1989a_wavelet}, well known as ``mathematical microscope'',
are effective time-frequency analysis tools,
which could be applied to decompose an image $\textbf{\emph{X}}$ into
the low-frequency component $\textbf{\emph{X}}_{ll}$ containing the image main information
and the high-frequency components $\textbf{\emph{X}}_{lh}, \textbf{\emph{X}}_{hl}, \textbf{\emph{X}}_{hh}$
containing the details (Fig. \ref{fig_MaxPooling_DWT_IDWT}).
In this paper, we rewrite Discrete Wavelet Transform (DWT) and Inverse DWT (IDWT) as the general network layers,
which are applicable to 1D/2D/3D data and various wavelets.
One can flexibly design end-to-end architectures using them,
and directly process the data details in the deep networks.
We design wavelet integrated deep networks for image segmentation, termed WaveSNets,
by replacing the down-sampling with DWT and the up-sampling with IDWT
in the U-Net \cite{ronneberger2015u_net}, SegNet \cite{badrinarayanan2017segnet}, and DeepLabv3+ \cite{chen2018encoder_deeplabv3+}.
When Haar, Cohen, and Daubechies wavelets are used,
we evaluate WaveSNets using dataset
CamVid \cite{brostow2009semantic_CamVid}, Pascal VOC \cite{everingham2015pascal}, and Cityscapes \cite{cordts2016cityscapes}.
The experimental results show that WaveSNets achieve better performances in semantic image segmentation than their vanilla versions,
due to the effective segmentation for the fine and similar objects.
In summary:
\begin{itemize}
\item We rewrite DWT and IDWT as general network layers, which are applicable to various wavelets
and can be applied to design end-to-end deep networks for processing the data details during the network inference;
\item We design WaveSNets using various network architectures,
by replacing the down-sampling with DWT layer and up-sampling with IDWT layer;
\item WaveSNets are evaluated on the dataset of CamVid, Pascal VOC, Cityscapes, and achieve better performance for semantic image segmentation.
\end{itemize}

\section{Related works}
\subsection{Sampling operation}
Down-sampling operations, such as max-pooling, average-pooling, and strided-convolution,
are introduced into the deep networks for local connectivity and weight sharing.
These down-sampling operations usually ignore the chassic sampling theorem \cite{nyquist1928certain},
which result in aliasing among the data components in different frequency intervals.
As a result, the data details presented by the high-frequency components are totally lost,
and random noises showing up in the same components could be sampled into the low resolution data.
In addition, the object basic structures presented by the low-frequency component will be broken.
Fig. \ref{fig_MaxPooling_DWT_IDWT} shows a max-pooling example.
In the signal processing, the low-pass filtering before the down-sampling is the standard method for anti-aliasing.
Anti-aliased CNNs \cite{zhang2019making} integrate the low-pass filtering with the down-sampling in the deep networks,
which achieve increased classification acccuracy and better shift-robustness.
However, the filters used in anti-aliased CNNs are empirically designed based on the row vectors of Pascal's triangle,
which are ad hoc and no theoretical justifications are given.
As no up-sampling operation, i.e., reconstruction, of the low-pass filter is available,
the anti-aliased U-Net \cite{zhang2019making} has to apply the same filtering after normal up-sampling to achieve the anti-aliasing effect.

Up-sampling operations, such as max-unpooling \cite{badrinarayanan2017segnet},
deconvolution \cite{ronneberger2015u_net}, and bilinear interpolation \cite{chen2014semantic_deeplabv1,chen2018encoder_deeplabv3+},
are widely used in the deep networks for image-to-image translation tasks.
These up-sampling operations are usually applied to gradually recover the data resolution,
while the data details can not be recovered from them.
Fig. \ref{fig_MaxPooling_DWT_IDWT} shows a max-unpooling example.
The lost data details would significantly degrade the network performance for the image-to-image tasks, such as the image segmentation.
Various techniques,
including \`atrous convolution \cite{chen2014semantic_deeplabv1,chen2018encoder_deeplabv3+}, PointRend \cite{kirillov2019pointrend}, etc.,
are introduced into the design of deep networks to capture the fine details for performance improvement of image segmentation.
However, these techniques try to recover the data details from the detail-unrelated information.
Their ability in the improvement of segmentation performance is limited.

\subsection{Wavelet}
Wavelets are powerful time-frequency analysis tools,
which have been widely used in signal analysis, image processing, and pattern recognition.
A wavelet is usually associated with scaling function and wavelet functions.
The shifts and expansions of these functions compose stable basis for the signal space,
with which the signal can be decomposed and reconstructed.
The functions are closely related to the low-pass and high-pass filters of the wavelet.
In practice, these filters are applied for the data decomposition and reconstruction.
As Fig. \ref{fig_MaxPooling_DWT_IDWT} shows,
2D Discrete Wavelet Transform (DWT) decompose the image $\textbf{\emph{X}}$ into its low-frequency component $\textbf{\emph{X}}_{ll}$
and three high-frequency components $\textbf{\emph{X}}_{lh}, \textbf{\emph{X}}_{hl}, \textbf{\emph{X}}_{hh}$.
While $\textbf{\emph{X}}_{ll}$ is a low resolution version of the image, keeping its main information,
$\textbf{\emph{X}}_{lh}, \textbf{\emph{X}}_{hl}, \textbf{\emph{X}}_{hh}$ save its horizontal, vertical, and diagonal details, respectively.
2D Inverse DWT (IDWT) could reconstruct the image using the DWT output.

Various wavelets,
including orthogonal wavelets, biorthogonal wavelets, multiwavelets, ridgelet, curvelet, bandelet and contourlet, etc.,
have been designed, studied, and applied in signal processing, numerical analysis, pattern recognition,
computer vision and quantum mechanics, etc.
The \`atrous convolution used in DeepLab \cite{chen2014semantic_deeplabv1,chen2018encoder_deeplabv3+}
is originally developed in the wavelet theory.
In the deep learning, while wavelets are widely applied as data preprocessing or postprocessing tools,
wavelet transforms are also introduced into the design of deep networks by taking them as substitutes of sampling operations.

Multi-level Wavelet CNN (MWCNN) \cite{liu2019multi_MWCNN}
integrates Wavelet Package Transform (WPT) into the deep network for image restoration.
MWCNN concatenates the low-frequency and high-frequency components of the input feature map, and processes them in a unified way.
The details stored in the high-frequency components would be largely wiped out via this processing mode,
because the data amplitude in the components is significantly weaker than that in the low-frequency component.
Convolutional-Wavelet Neural Network (CWNN) \cite{duan2017sar} applies the redundant dual-tree complex wavelet transform (DT-CWT)
to suppress the noise and keep the object structures for extracting robust features from SAR images.
The architecture of CWNN contains only two convolution layers.
While DT-CWT discards the high-frequency components output from DT-CWT,
CWNN takes as its down-sampling output the average value of the multiple low-frequency components.
Wavelet pooling proposed in \cite{williams2018wavelet} is designed using a two-level DWT.
Its back-propagation performs a one-level DWT with a two-level IDWT,
which does not follow the mathematical principle of gradient.
Recently, the application of wavelet transform in image style transfer \cite{yoo2019photorealistic} is studied.
In these works, the authors evaluate their methods using only one or two wavelets,
because of the absence of the general wavelet transform layers;
the data details presented by the high-frequency components are abandoned or processed together with the low-frequency component,
which limits the detail restoration in the image-to-image translation tasks.

\section{Our method}
Our method is to replace the sampling operations in the deep networks with wavelet transforms.
We firstly rewrite Discrete Wavelet Transform (DWT) and Inverse DWT (IDWT) as the general network layers.
Although the following analysis is mainly for orthogonal wavelet and 1D data,
it can be generalized to other wavelets and 2D/3D data with slight changes.

\subsection{Wavelet transform}
For a given 1D data $\textbf{\emph{x}} \in \mathbb{R}^n$,
DWT decomposes it, using two filters, i.e., low-pass filter $\textrm{f}_l$ and high-pass filter $\textrm{f}_h$ of an 1D orthogonal wavelet,
into its low-frequency component $\textbf{\emph{x}}_l$
and high-frequency component $\textbf{\emph{x}}_h$,
where
\begin{equation}\label{eq_DWT_1D}
\textbf{\emph{x}}_c = (\downarrow2)(\textrm{f}_c \ast \textbf{\emph{x}}), \quad c \in \{l,h\},
\end{equation}
and $\ast$, $(\downarrow2)$ denote the convolution and naive down-sampling, respectively.
In theory, the length of every component is $1/2$ of that of $\textbf{\emph{x}}$, i.e.,
\begin{equation}
\textbf{\emph{x}}_c \in \mathbb{R}^{\lfloor \frac{n}{2} \rfloor}, \quad c \in \{l,h\}.
\end{equation}
Therefore, $n$ is usually even number.

IDWT reconstructs the original data $\textbf{\emph{x}}$ based on the two components,
\begin{equation}\label{eq_IDWT_1D}
\textbf{\emph{x}} = \textrm{f}_l \ast (\uparrow2)\,\textbf{\emph{x}}_l + \textrm{f}_h \ast (\uparrow2)\,\textbf{\emph{x}}_h,
\end{equation}
where $(\uparrow2)$ denotes the naive up-sampling operation.

For high-dimensional data, high dimensional DWT could decomposes it into one low-frequency component and multiple high-frequency components,
while the corresponding IDWT could reconstructs the original data from the DWT output.
For example, for a given 2D data $\textbf{\emph{X}} \in \mathbb{R}^{m\times n}$,
2D DWT decomposes it into four components,
\begin{equation}\label{eq_DWT_2D}
\textbf{\emph{X}}_{c_0c_1} = (\downarrow2)(\textrm{f}_{c_0c_1} \ast \textbf{\emph{X}}), \quad c_0, c_1 \in \{l,h\},
\end{equation}
where $\textrm{f}_{ll}$ is the low-pass filter and $\textrm{f}_{lh}, \textrm{f}_{hl}, \textrm{f}_{hh}$ are the high-pass filters of the 2D orthogonal wavelet.
$\textbf{\emph{X}}_{ll}$ is the low-frequency component of the original data which is a low-resolution version containing the data main information;
$\textbf{\emph{X}}_{lh}, \textbf{\emph{X}}_{hl}, \textbf{\emph{X}}_{hh}$ are three high-frequency components which store the vertical, horizontal, and diagonal details of the data.
2D IDWT reconstruct the original data from these components,
\begin{equation}\label{eq_IDWT_2D}
\textbf{\emph{X}} = \sum_{c_0, c_1 \in \{l,h\}}\textrm{f}_{c_0c_1} \ast (\uparrow2)\,\textbf{\emph{X}}_{c_0c_1}.
\end{equation}
Similarly, the size of every component is $1/2$ size of the original 2D data in terms of the two dimensional direction, i.e.,
\begin{equation}
\textbf{\emph{X}}_{c_0c_1} \in \mathbb{R}^{\lfloor\frac{m}{2}\rfloor\times\lfloor\frac{n}{2}\rfloor},\quad c_0,c_1\in\{l,h\}.
\end{equation}
Therefore, $m,n$ are usually even numbers.

Generally, the filters of high-dimensional wavelet are tensor products of the two filters of 1D wavelet.
For 2D wavelet, the four filters could be designed from
\begin{equation}\label{eq_filter_2D}
\textrm{f}_{c_0c_1} = \textrm{f}_{c_1} \otimes \textrm{f}_{c_0}, \quad c_0, c_1 \in \{l,h\},
\end{equation}
where $\otimes$ is the tensor product operation.
For example, the low-pass and high-pass filters of 1D Haar wavelet are
\begin{equation}\label{eq_filter_2D}
\textrm{f}_l^{\textrm{H}} = \frac{1}{\sqrt{2}} (1,1)^T, \quad
\textrm{f}_h^{\textrm{H}} = \frac{1}{\sqrt{2}} (1,-1)^T.
\end{equation}
Then, the filters of the corresponding 2D Haar wavelet are
\begin{align}
\textrm{f}_{ll}^{\textrm{H}} &= \textrm{f}_l^{\textrm{H}} \otimes \textrm{f}_l^{\textrm{H}} =
\frac{1}{2}\left[
\begin{array}{cc}
1&1\\1&1
\end{array}\right]\\
\textrm{f}_{hl}^{\textrm{H}} &= \textrm{f}_l^{\textrm{H}} \otimes \textrm{f}_h^{\textrm{H}} =
\frac{1}{2}\left[
\begin{array}{cc}
1&-1\\1&-1
\end{array}\right]\\
\textrm{f}_{lh}^{\textrm{H}} &= \textrm{f}_h^{\textrm{H}} \otimes \textrm{f}_l^{\textrm{H}} =
\frac{1}{2}\left[
\begin{array}{cc}
1&1\\-1&-1
\end{array}\right]\\
\textrm{f}_{hh}^{\textrm{H}} &= \textrm{f}_h^{\textrm{H}} \otimes \textrm{f}_h^{\textrm{H}} =
 \frac{1}{2}\left[
\begin{array}{cc}
1&-1\\-1&1
\end{array}\right]
\end{align}

Eqs. (\ref{eq_DWT_1D}), (\ref{eq_IDWT_1D}), (\ref{eq_DWT_2D}) and (\ref{eq_IDWT_2D}) present the forward propagations for 1D/2D DWT and IDWT.
It is onerous to deduce the gradient for the backward propagations from these equations.
Fortunately, the modern deep learning framework PyTorch \cite{paszke2017automatic} could automatically deduce the gradients for the common tensor operations.
We have rewrote 1D/2D DWT and IDWT as network layers in PyTorch,
which will be publicly available for other researchers.
In the layers, we do DWT and IDWT channel by channel for multi-channel data.

\subsection{WaveSNet}
We design wavelet integrated deep networks for image segmentation (WaveSNets),
by replacing the down-sampling operations with 2D DWT and the up-sampling operations with 2D IDWT.
In this paper, we take U-Net, SegNet, and DeepLabv3+ as the basic architectures.

\begin{figure}[bpt]
	\centering
	\subfigure[PDDS]{
		\label{fig_dual_structure_Pooling_Deconv}
		\includegraphics*[scale=0.525, viewport=72 685 285 767]{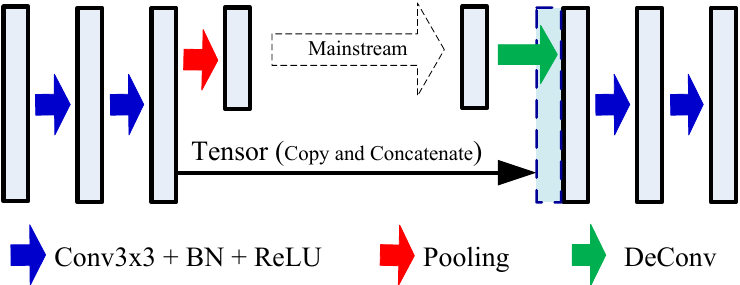}
	}\hspace{7pt}
	\subfigure[PUDS]{
		\label{fig_dual_structure_Pooling_Unpooling}
		\includegraphics*[scale=0.525, viewport=72 479 285 561]{figures/dual_structures.pdf}
	}
	\caption{The dual structures used in U-Net and SegNet.}\label{fig_dual_structures}
\end{figure}
\begin{figure}[bpt]
	\centering
	\includegraphics*[scale=0.7, viewport=72 380 285 462]{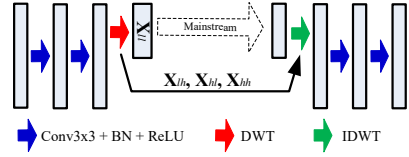}
	\caption{The wavelet based dual structure (WADS).}\label{fig_dual_structures_wavelet}
\end{figure}
\textbf{WSegNets}\quad
SegNet and U-Net share a similar symmetrical encoder-decoder architecture, but differ in their sampling operations.
We name the pair of connected down-sampling and up-sampling operations and the associated convolutional blocks as \textbf{dual structure},
where the convolutional blocks process the feature maps with the same size.
Fig. \ref{fig_dual_structure_Pooling_Deconv} and Fig. \ref{fig_dual_structure_Pooling_Unpooling}
show the dual structures used in U-Net and SegNet,
which are named as PDDS (Pooling Deconvolution Dual Structure) and PUDS (Pooling-Unpooling Dual Structure), respectively.
U-Net and SegNet consist of multiple nested dual structures.
While they apply the max-pooling during their down-sampling,
PDDS and PUDS use deconvolution and max-unpooling for the up-sampling, respectively.
As Fig. \ref{fig_dual_structure_Pooling_Deconv} shows, PDDS copys and transmits the feature maps from encoder to decoder,
concatenating them with the up-sampled features and extracting detailed information for the object boundaries restoration.
However, the data tensor injected to the decoder might contain redundant information,
which interferes with the segmentation results and introduces more convolutional paramters.
PUDS transmits the pooling indices via the branch path for the upgrading of the feature map resolution in the decoder.
As Fig. \ref{fig_MaxPooling_DWT_IDWT} shows, the lost data details can not be restored from the pooling indices.

To overcome the weaknesses of PDDS and PUDS, we adopt DWT for down-sampling and IDWT for up-sampling,
and design WADS (WAvelet Dual Structure, Fig. \ref{fig_dual_structures_wavelet}).
During its down-samping, WADS decomposes the feature map into low-frequency and high-requency components.
Then, WADS injects the low-frequency component into the following layers in the deep networks to extract high-level features,
and transmits the high-frequency components to the up-sampling layer for the recovering of the feature map resolution using IDWT.
IDWT could also restore the data details from the high-frequency components during the up-sampling.
We design wavelet integrated encoder-decoder networks using WADS, termed WSegNets, for imag segmentation.

Table \ref{Tab_network_configuration} illustrates the configuration of WSegNets, together with that of U-Net and SegNet.
In this paper, the U-Net consists of eight more convolutional layers than the original one \cite{ronneberger2015u_net}.
\begin{table}[!t]
	\scriptsize
	\begin{center}
	\setlength{\tabcolsep}{1.5mm}{
	\begin{tabular}{c||l|r||c|c|c}\hline
	\multirow{3}{*}{data size} &    \multicolumn{2}{c||}{the number of channels}     & \multicolumn{3}{c}{networks}  \\ \cline{2-6}
	& \multicolumn{1}{c|}{encoder} & \multicolumn{1}{c||}{decoder} &SegNet & U-Net & WSegNet       \\ \hline
				      $512\times512$       & 3, 64                    &                   64, 64 &          PUDS           &          PDDS           &  WADS         \\
				      $256\times256$       & 64, 128                  &                  128, 64 &          PUDS           &          PDDS           &  WADS         \\
				      $128\times128$       & 128, 256, 256            &            256, 256, 128 &          PUDS           &          PDDS           &  WADS         \\
				       $64\times64$        & 256, 512, 512            &            512, 512, 256 &          PUDS           &          PDDS           &  WADS         \\
				       $32\times32$        & 512, 512, 512            &            512, 512, 512 &          PUDS           &          PDDS           &  WADS         \\ \hline
\end{tabular}}
\caption{Configurations of U-Net, SegNet, and WSegNet.} \label{Tab_network_configuration}
\end{center}
\end{table}
In Table \ref{Tab_network_configuration}, the first column shows the input size,
though these networks can process images with arbitrary size.
Every number in the table corresponds to a convolutional layer followed by a Batch Normalization (BN) and Rectified Linear Unit (ReLU).
While the number in the column ``encoder'' is the number of the input channels of the convolution,
the number in the column ``decoder'' is the number of the output channels.
The encoder of U-Net and SegNet consists of 13 convolutional layers
corresponding to the first 13 layers in the VGG16bn \cite{simonyan2014very_VGG}.
A convolutional layer with kernel size of $1\times1$ converts the decoder output into the predicted segmentation result.

\textbf{WDeepLabv3+}\quad
DeepLabv3+ \cite{chen2018encoder_deeplabv3+} employs an unbalanced encoder-decoder architecture.
The encoder applys an \`atrous convolutional version of CNN
to alleviate the detailed information loss due to the common down-sampling operations,
and an \`Atrous Spatial Pyramid Pooling (ASPP) to extract image multiscale representations.
At the begin of the decoder, the encoder feature map is directly upsampled using a bilinear interpolation with factor of 4,
and then concatenated with a low-level feature map transmitted from the encoder.
DeepLabv3+ adopts a dual structure connecting its encoder and decoder,
which differs with PDDS only on the up-sampling and suffers the same drawbacks.

\begin{figure}[bpt]
	\centering
	\includegraphics*[scale=0.475, viewport=27 473 534 713]{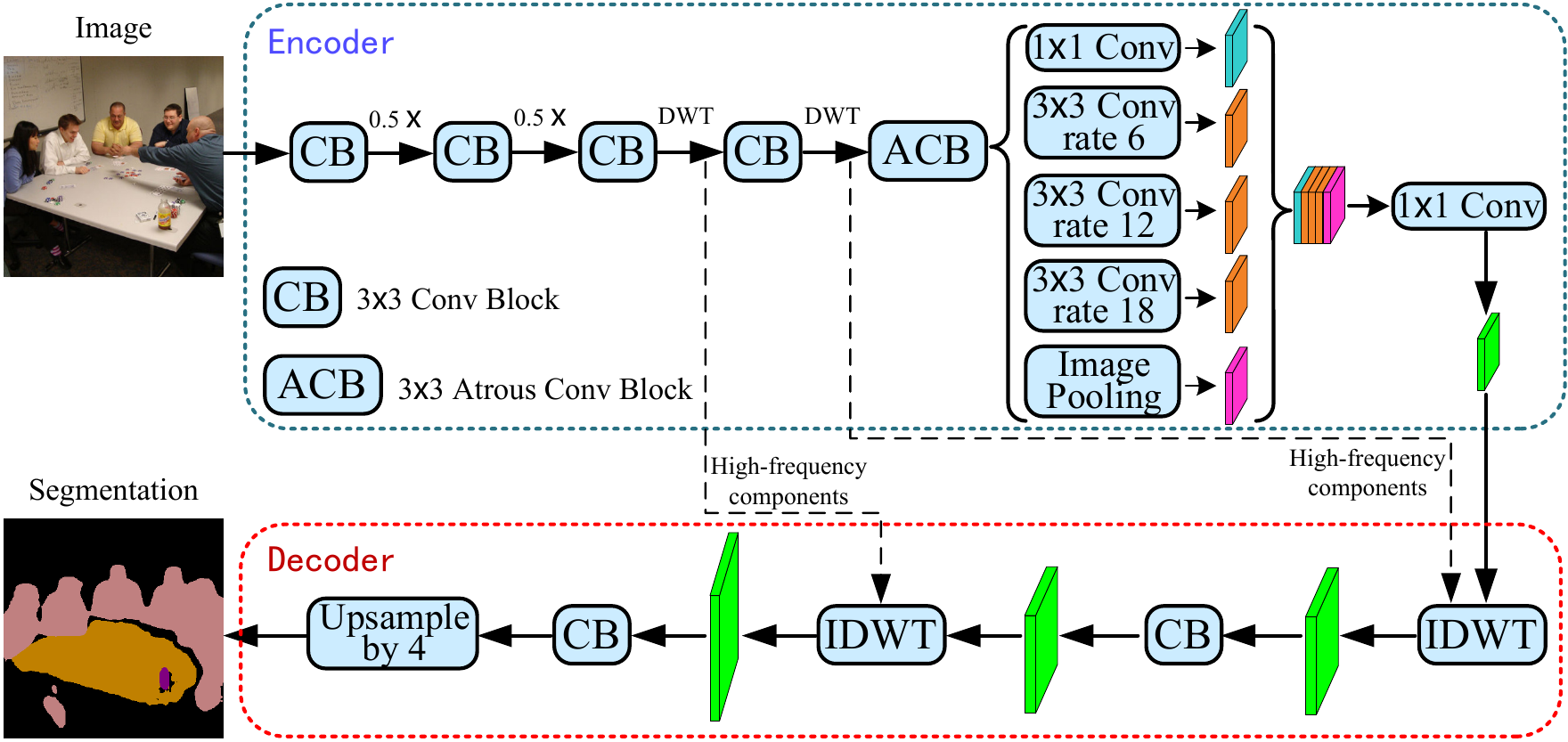}
	\caption{WDeepLabv3+ structure.}
	\label{fig_WDeepLabv3+}
\end{figure}
We design WDeepLabv3+, a wavelet integrated version of DeepLabv3+,
by applying two wavelet version of dual structures.
As Fig. \ref{fig_WDeepLabv3+} shows, the encoder applys a wavelet integrated \`atrous CNN followed by the ASPP,
which output encoder feature map and two sets of high-frequency components.
The encoder feature map is up-sampled by two IDWT, integrated with the detail information contained in the high-frequency components,
while the two IDWT are connected with a convolutional block.
The decoder then apply a few $3\times3$ convolutions to refine the feature map,
followed by a bilinear interpolation to recover the resolution.

\section{Experiment}
We evaluate the WaveSNets (WSegNet, WDeepLabv3+) on the image dataset of
CamVid, Pascal VOC, and Cityscapes, in terms of mean intersection over union (mIoU).

\subsection{WSegNet}
\textbf{CamVid}\quad
CamVid contains 701 road scene images (367, 101 and 233 for the training, validation and test respectively) with size of $360\times480$,
which was used in \cite{badrinarayanan2017segnet} to quantify SegNet performance.
Using the training set, we train SegNet, U-Net, and WSegNet when various wavelets are applied.
The trainings are supervised by cross-entropy loss.
We employ SGD solver,
initial learning rate of $0.007$ with ``poly'' policy and $power = 0.9$, momentum of $0.9$ and weight decay of $0.0005$.
With a mini-batch size of $20$, we train the networks 12K iterations (about 654 epochs).
The input image size is $352 \times 480$.
For every image, we adopt random resizing between $0.75$ and $2$,
random rotation between $-20$ and $20$ degrees, random horizontal flipping and cropping with size of $352 \times 480$.
We apply a pre-trained VGG16bn model for the encoder initialization
and initialize the decoder using the technique proposed in \cite{he2015delving_initilization}.
We do not tune above hyper-parameters for a fair comparison.

\begin{figure}[bpt]
	\centering
	\includegraphics*[scale=0.55, viewport=41 430 474 570]{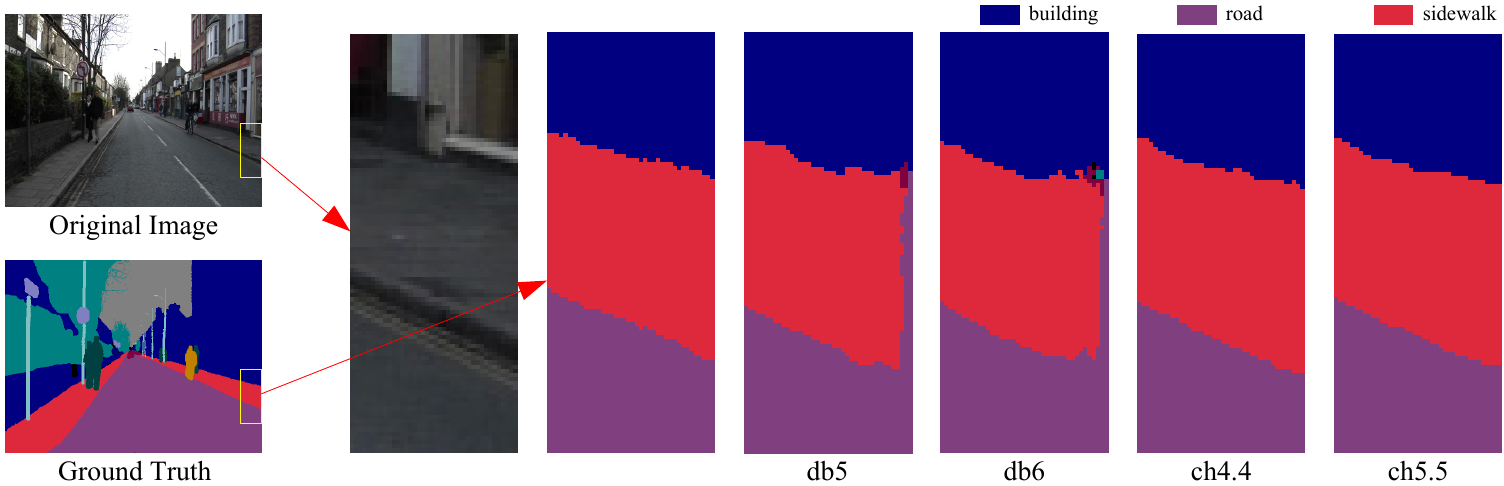}
	\caption{Boundary comparison for images segmented by WSegNet with different wavelets.}
	\label{Fig_CamVid_Comparison_Boundary}
\end{figure}

\begin{table}[!tbp]
\scriptsize
\centering
\setlength{\tabcolsep}{0.35mm}{
\begin{tabular}{r|c|c|c|cccc|ccccc}\hline
&\multirow{2}{*}{SegNet}&\multirow{2}{*}{U-Net}&\multicolumn{10}{c}{WSegNet}\\\cline{4-13}
&&&haar&ch2.2&ch3.3&ch4.4&ch5.5&db2&db3&db4&db5&db6\\\hline
sky        &90.50    &\textbf{91.52}   &91.31  &91.38  &91.29  &91.39  &91.24  &91.35 &91.48    &90.99       &90.89       &90.63\\
building   &77.80    &\textbf{80.28}   &79.90  &79.27  &78.82  &79.37  &78.65  &79.48 &79.60    &78.52       &78.58       &77.84\\
pole       &~~9.14   &27.97   &27.99  &\textbf{29.38}  &27.90  &28.96  &27.26  &27.91 &28.38    &28.04       &26.66       &25.96\\
road       &92.69    &93.71   &93.69  &93.47  &93.47  &93.77  &93.78  &93.72 &\textbf{93.91}    &92.57       &92.12       &92.11\\
sidewalk   &78.05    &80.05   &80.33  &79.44  &79.34  &79.89  &80.08  &79.67 &\textbf{80.58}    &76.95       &75.62       &76.65\\
tree       &72.40    &73.51   &73.34  &73.27  &73.21  &73.07  &71.60  &73.61 &\textbf{73.68}    &72.90       &72.28       &71.92\\
symbol     &37.61    &43.07   &44.44  &42.68  &40.42  &43.57  &42.33  &\textbf{44.72} &44.01    &41.06       &41.72       &39.69\\
fence      &19.92    &27.50   &\textbf{32.59}  &24.62  &25.59  &28.40  &28.85  &25.52 &29.60    &26.90       &24.15       &29.00\\
car        &79.31    &\textbf{85.04}   &83.21  &84.43  &82.63  &84.57  &84.14  &84.30 &83.97    &81.92       &81.07       &78.38\\
walker     &39.93    &50.35   &49.35  &\textbf{50.98}  &50.52  &50.43  &49.09  &50.15 &49.39    &47.69       &48.02       &43.17\\
bicyclist  &39.48    &44.45   &50.38  &47.94  &48.69  &47.93  &\textbf{52.64}  &51.15 &47.73    &46.08       &43.53       &38.96\\\cdashline{1-13}[3pt/1pt]
mIoU       &57.89    &63.40   &\textbf{64.23}  &63.35  &62.90  &63.76  &63.61  &63.78 &63.85    &62.15       &61.33       &60.39\\\hline
gl. acc.    &91.04	 &92.19   &92.08	&92.04	&91.98	&92.10	&91.73	&92.06 &\textbf{92.40}  &91.60      &91.30       &91.14\\\hline
para.(e6)   &29.52   &37.42&\multicolumn{10}{c}{\textbf{29.52}}\\\hline
\end{tabular}}
\caption{WSegNet results on CamVid test set}\label{tab_mIoU_CamVid_WSegNet}
\end{table}
Table \ref{tab_mIoU_CamVid_WSegNet} shows the mIoU and global accuracy on the CamVid test set,
together with the parameter numbers of U-Net, SegNet, and WSegNet with various wavelets.
In the table, ``dbx'' represents orthogonal Daubechies wavelet with approximation order $x$,
and ``chx.y'' represents biorthogonal Cohen wavelet with approximation orders $(x,y)$.
Their low-pass and high-pass filters can be found in \cite{daubechies1992ten}.
The length of these filters increase as the order increases.
While Haar and Cohen wavelets are symmetric, Daubechies are not.
The mIoU of WSegNet decreases from $63.78\%$ to $60.39\%$ as asymmetric wavelet varies from ``db2'' to ``db6'',
while it varies from $63.35\%$ to $63.61\%$ as symmetric wavelet varies from ``ch2.2'' to ``ch5.5''.
It seems that the performances among different asymmetric wavelets are much diverse than that among various symmetric wavelets.
In the wavelet integrated deep networks, we truncate the DWT output to make them to be $1/2$ size of input data.
As a result, IDWT with asymmetric wavelet can not fully restore an image in the region near the image boundary,
and the region width increases as the wavelet filter length increases.
With symmetric wavelet, however, one can fully restore the image based on symmetric extension of the DWT output.
Consequently, WSegNet with Cohen wavelets performs better than that with Daubechies ones near the image boundary.
Fig. \ref{Fig_CamVid_Comparison_Boundary} shows an example image, its manual annotation,
a region consisting of ``building'', ``road'' and ``sidewalk'' and the segmentation results achieved using different wavelets.
The region is located close to the image boundary and has been enlarged with colored segmentation results for better illustration.
One can observe from the results of ``db5'' and ``db6'' that a long line of ``sidewalk'' pixels,
located near the image boundary, are classified as ``road''.
In comparison, the results of ``ch4.4'' and ``ch5.5'' match well with the ground truth.

In \cite{badrinarayanan2017segnet}, the authors train the SegNet using an extended CamVid training set containing 3433 images,
which achieved $60.10\%$ mIoU on the CamVid test set.
We train SegNet, U-Net and WSegNet using only 367 CamVid training images.
From Table \ref{tab_mIoU_CamVid_WSegNet}, one can find WSegNet achieves the best mIoU ($64.23\%$)  using Haar wavelet,
and the best global accuracy ($92.40\%$) using ``db3'' wavelet.
WSegNet is significantly superior to SegNet in terms of mIoU and the global accuracy,
while they require the same amount of parameters ($29.52\times10^6$).
The segmentation performance of WSegNet is also better than that of U-Net,
while it requires much less parameters than U-Net ($37.42\times10^6$).

Table \ref{tab_mIoU_CamVid_WSegNet} also lists the IoUs for the 11 categories in the CamVid.
Compared with U-Net and WSegNet, SegNet performs very poor on the fine objects, such as ``pole'', ``symbol'', and ``fence'',
as that the max-pooling indices used in SegNet are not helpful for restoring the image details.
While U-Net achieves comparable or even better IoUs on the easily identifiable objects, such as the ``sky'', ``building'', and ``car'',
it does not discriminate well similar objects like``walker'' and ``bicyclist'', or ``building'' and ``fence''.
The feature maps of these two pairs of objects might look similar to the decoder of U-Net, as the data details are not separately provided.
Table \ref{tab_confusion_matrices} shows the confusion matrices on these four categories.
The proportion of ``building'' in the predicted ``fence''
decreases from $34.11\%$ to $30.16\%$, as the network varies from SegNet to WPUNet,
while that of ``bicyclist'' in the predicted ``walker'' decreases from $2.54\%$ to $1.90\%$.
These results suggest that WSegNet is more powerful than SegNet and U-Net in distinguishing similar objects.
\begin{table}
	\scriptsize
	\centering
		\setlength{\tabcolsep}{1.25mm}{
			\begin{tabular}{r|cc|cc|cc}
				\hline
				& \multicolumn{2}{|c|}{SegNet}          &\multicolumn{2}{c|}{U-Net}       &\multicolumn{2}{c}{WSegNet(haar)}\\\cline{2-7}
				&        building   &        fence      &        building   &        fence      & building          &        fence\\ \hline
	   building &       88.01       &      ~~1.30       &       89.91       &      ~~1.19       &       90.22       &      ~~1.08\\
		  fence & \underline{34.11} &       29.12       & \underline{30.64} &       40.16       & \underline{30.16} &       44.65\\ \hline\hline
				&        walker&        bicyclist&        walker&        bicyclist&        walker&        bicyclist\\ \hline
		 walker &       53.30       & \underline{~~2.54}  &       66.83       & \underline{~~1.95}&       68.74       & \underline{~~1.90}\\
	  bicyclist &       13.69       &        49.66        &       16.73       &       51.45       &       12.52       &       59.80 \\ \hline
		\end{tabular}}
	\caption{Confusion matrices for the CamVid test set.}
	\label{tab_confusion_matrices}
\end{table}
\begin{figure}[bpt]
	\centering
	\includegraphics*[scale=0.675, viewport=40 616 382 776]{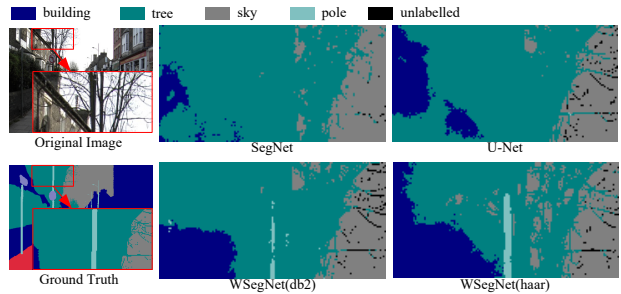}
	\caption{Comparison of SegNet and WSegNet segmentations.}
	\label{fig_segmentation_comparison_CamVid}
\end{figure}

Fig. \ref{fig_segmentation_comparison_CamVid} presents a visual example for various networks,
which shows the example image, its manual annotation,
a region consisting of ``building'', ``tree'', ``sky'' and ``pole'', and the segmentation results achieved using SegNet, U-Net and WSegNet.
The region is enlarged with colored segmentation reults for better illustration.
From the figure, one can find in the segmentation result that WSegNet keeps the basic structure of ``tree'',
``pole'', and ``building'' and restores the object details, such as the ``tree'' branches and the ``pole''.
The segmentation result of WSegNet matches the image region much better than that of SegNet and U-Net,
even corrects the annotation noises about ``building'' and “tree” in the ground truth.

\begin{table}[!tbp]
\scriptsize
\centering
\setlength{\tabcolsep}{0.75mm}{
\begin{tabular}{cr|c|c|c|cccc|c}\hline
&&\multirow{2}{*}{SegNet}&\multirow{2}{*}{U-Net}&\multicolumn{6}{c}{WSegNet}\\\cline{5-10}
	&					&					&		&	haar&	ch2.2&	ch3.3&	ch4.4&	ch5.5&	db2\\\hline
\multirow{2}{*}{Pascal VOC}&mIoU     &61.33 	&63.64	&\textbf{63.95}	&63.50 	&63.46	&63.48	&63.62	&63.48\\\cdashline{2-10}[1pt/1pt]
							&gl. acc.&90.14		&\textbf{90.82}	&90.79	&90.72	&90.72	&90.73	&90.76	&90.75\\\hline
\multirow{2}{*}{Cityscapes}&mIoU     &65.75		&70.05	&70.09	&69.86	&69.73	&70.13	&\textbf{70.63}	&70.37	\\\cdashline{2-10}[1pt/1pt]
							&gl. acc.&94.65		&\textbf{95.21}	&95.20	&95.18	&95.09	&95.17	&95.15	&95.20  \\\hline
\multicolumn{2}{r|}{parameters (e6)}& 29.52&37.42		&\multicolumn{6}{c}{\textbf{29.52}}\\\hline
\end{tabular}}
\caption{WSegNet results on Pascal VOC and Cityscapes \textit{val} set.}\label{tab_mIoU_VOC_Cityscapes_WSegNet}
\end{table}
\textbf{Pascal VOC and Cityscapes}\quad
The original Pascal VOC2012 semantic segmentation dataset contains 1464 and 1449 annotated images
for training and validation, respectively, and contains 20 object categories and one background class.
The images are with various sizes.
We augment the training data to 10582 by extra annotations provided in \cite{hariharan2011semantic}.
We train SegNet, U-Net, and WSegNet with various wavelets on the extended training set 50 epochs with batch size of 16.
During the training, we adopt random horizontal flipping, random Gaussian blurring, and cropping with size of $512\times512$.
The other hyper-parameters are the same with that used in CamVid training.
Table \ref{tab_mIoU_VOC_Cityscapes_WSegNet} presents the results on the validation set for the trained networks.

Cityscapes contains 2975 and 500 high quality annotated images for the training and validation, respectively.
The images are with size of $1024\times2048$.
We train the networks on the training set 80 epochs with batch size of 8 and initial learning rate of $0.01$.
During the training, we adopt random horizontal flipping, random resizing between 0.5 and 2, and random cropping with size of $768\times768$.
Table \ref{tab_mIoU_VOC_Cityscapes_WSegNet} presents the results on the validation set.

From Tabel \ref{tab_mIoU_VOC_Cityscapes_WSegNet}, one can find the segmentation performance of SegNet
is significant inferior to that of U-Net and WSegNet.
While WSegNet achieves better mIoU ($63.95\%$ for Pascal VOC and $70.63\%$ for Cityscapes)
and requires less parameters ($29.52\times10^6$),
the global accuracies of WSegNet on the two dataset are a little lower than that of U-Net.
This result suggest that, compared with U-Net, WSegNet could more precisely classify the pixels at the ``critical'' locations.

\subsection{WDeepLabv3+}
\begin{table}[!tbp]
\scriptsize
\centering
\setlength{\tabcolsep}{1.25mm}{
\begin{tabular}{r|c|c|cccc|c}\hline
&\multirow{2}{*}{DeepLabv3+}&\multicolumn{6}{c}{WDeepLabv3+}\\\cline{3-8}
			&			&	haar&	ch2.2&	ch3.3&	ch4.4&	ch5.5&	db2\\\hline
background  &93.83         &93.82   &93.85      &\textbf{93.94}      &93.91      &93.86      &93.87 \\
aeroplane   &92.29         &93.14   &92.50      &91.56      &\textbf{93.21}      &92.73      &92.41 \\
bike        &41.42         &43.08   &42.19      &42.21      &\textbf{43.42}      &42.59      &42.84 \\
bird        &91.47         &90.47   &\textbf{91.60}      &90.34      &91.24      &90.81      &90.73 \\
boat        &75.39         &\textbf{75.47}   &72.04      &75.19      &74.20      &72.40      &72.90 \\
bottle      &82.05         &80.18   &81.14      &82.12      &79.55      &\textbf{82.23}      &81.89 \\
bus         &\textbf{93.64}         &93.25   &93.25      &93.20      &93.07      &93.52      &93.31 \\
car         &89.30         &90.36   &90.00      &\textbf{90.67}      &88.79      &86.31      &87.11 \\
cat         &93.69         &92.80   &93.31      &92.62      &93.56      &93.84      &\textbf{93.97} \\
\underline{chair}       &38.28         &40.80   &40.75      &39.32      &39.79      &\textbf{43.27}      &41.60 \\
\textbf{\emph{cow}}         &86.60         &89.72   &89.04      &90.49      &88.42      &\textbf{92.04}      &88.17 \\
\underline{table}       &61.37         &63.24   &62.67      &65.49      &64.93      &\textbf{67.31}      &65.58 \\
dog         &\textbf{91.16}         &90.24   &91.04      &89.54      &89.97      &90.38      &90.65 \\
\textbf{\emph{horse}}       &86.60         &88.86   &88.88      &90.23      &89.00      &\textbf{91.02}      &89.19 \\
motorbike   &88.47         &87.94   &87.89      &\textbf{88.61}      &88.36      &87.30      &87.58 \\
person      &86.71         &86.88   &86.61      &86.35      &86.59      &86.32      &\textbf{86.96} \\
\underline{plant}       &64.48         &64.33   &64.69      &\textbf{68.45}      &65.50      &68.01      &65.41 \\
\textbf{\emph{sheep}}       &83.04         &87.45   &86.50      &87.43      &85.70      &\textbf{88.14}      &85.95 \\
sofa        &49.24         &47.43   &48.06      &49.85      &46.74      &\textbf{51.94}      &50.19 \\
train       &85.49         &84.18   &83.88      &85.47      &83.88      &85.16      &\textbf{86.69} \\
monitor     &77.80         &76.55   &78.31      &79.04      &78.32      &78.78      &\textbf{79.11} \\\cdashline{1-8}[3pt/1pt]
mIoU		&78.68		   &79.06	&78.96		&79.62		&78.96		&\textbf{79.90}		&79.34 \\\hline
gl. acc. &94.64		   &94.69	&94.68		&94.75		&94.68		&\textbf{94.77}		&94.75 \\\hline	
para. (e6)  &\textbf{59.34}		   &\multicolumn{6}{c}{60.22}	\\\hline
\end{tabular}}
\caption{WDeepLabv3+ results on Pascal VOC \textit{val} set.}\label{tab_mIoU_VOC_WDeepLabv3+}
\end{table}
Taking ResNet101 as the backbone, we build DeepLabv3+ and WDeepLabv3+ with $output\ stride = 16$,
and train them on the Pascal VOC using the same training policy with that used in the training of WSegNet.
Table \ref{tab_mIoU_VOC_WDeepLabv3+} shows the segmentation results on the validation set.

We achieve $78.68\%$ mIoU using DeepLabv3+ on the Pascal VOC validation set,
which is comparable to that ($78.85\%$) obtained by the inventors of this network \cite{chen2018encoder_deeplabv3+}.
With a few increase of parameter ($0.88 \times 10^6,~1.48\%$),
the segmentation performance of WDeepLabv3+ with various wavelet is always better than that of DeepLabv3+,
in terms of mIoU and global accuracy.
Using ``ch5.5'' wavelet, WDeepLabv3+ achieves the best performance, $79.90\%$ mIoU and $94.77\%$ global accuracy.
From Table \ref{tab_mIoU_VOC_WDeepLabv3+}, one can find that the better performance of WDeepLabv3+
is mainly resulted from its better segmentation
for the fine objects (``chair'', ``table'', and ``plant'') and similar objects (``cow'', ``horse'', and ``sheep'').
The above results justify the high efficiency of DWT/IDWT layers in processing data details.

\begin{figure}[bpt]
	\centering
	\includegraphics*[scale=0.415, viewport=22 260 581 810]{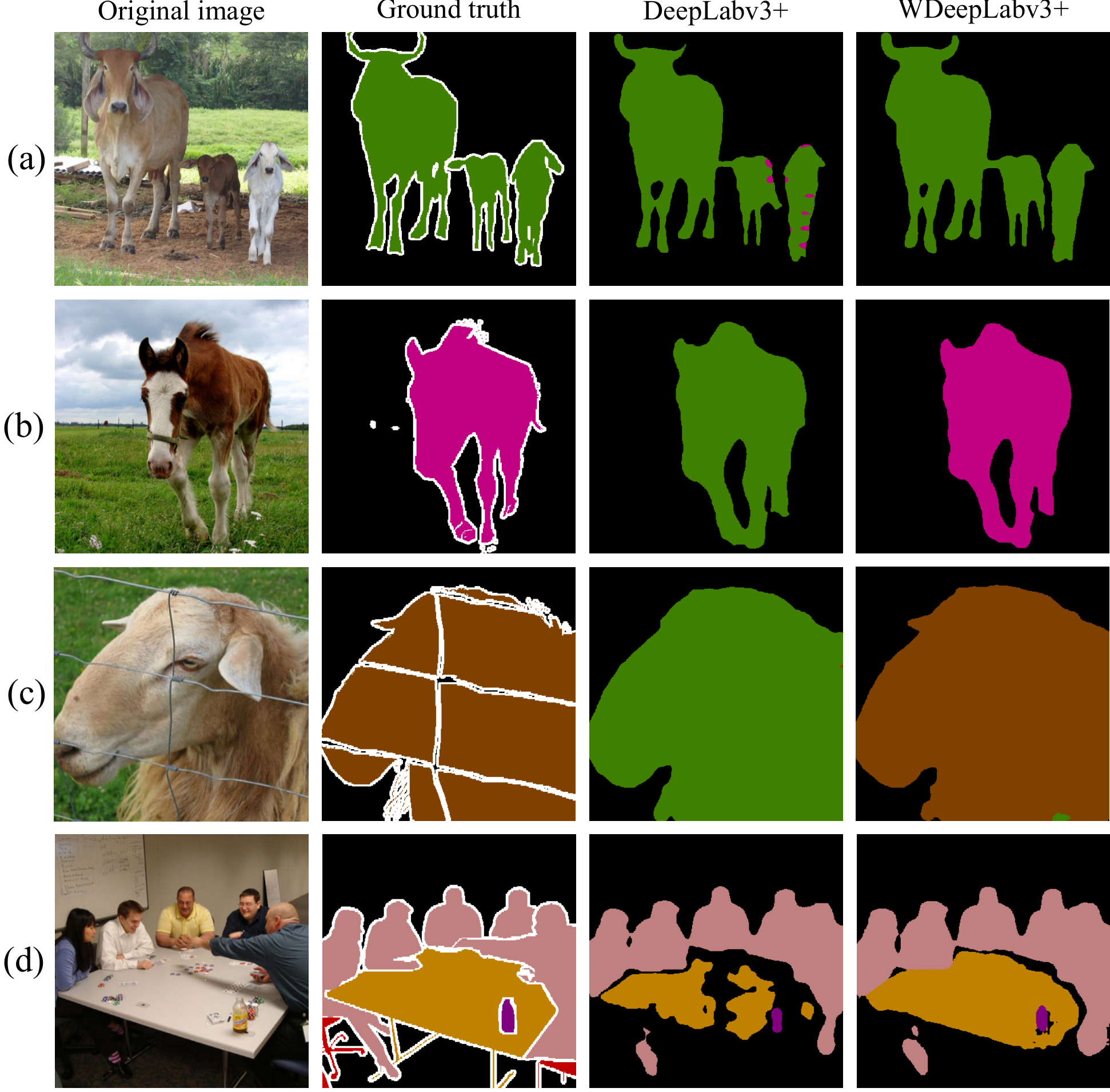}
	\caption{Comparison of DeepLabv3+ and WDeepLabv3+ results.}
	\label{fig_comparison_WDeepLabv3+_VOC}
\end{figure}
Fig. \ref{fig_comparison_WDeepLabv3+_VOC} shows four visual examples of segmentation results for DeepLabv3+ and WDeepLabv3+.
The first and second columns present the original images and the segmentation ground truth,
while the third and fourth columns show the segmentation results of DeepLabv3+ and WDeepLabv3+ with ``ch5.5'' wavelet, respectively.
We show the segmentation results with colored pictures for better illustration.
For the example image in Fig. \ref{fig_comparison_WDeepLabv3+_VOC}(a),
DeepLabv3+ falsely classifies the pixels in some detail regions on the cow and her calfs as ``background'' or ``horse'',
i.e., the regions for the cow's left horn, the hind leg of the brown calf, and some fine regions on the two calfs.
While WDeepLabv3+ correctly classifies the horse and the sheep
in the Fig. \ref{fig_comparison_WDeepLabv3+_VOC}(b) and Fig. \ref{fig_comparison_WDeepLabv3+_VOC}(c),
DeepLabv3+ classifies them as ``cow'' because of the similar object structures.
In Fig. \ref{fig_comparison_WDeepLabv3+_VOC}(d), the ``table'' segmented by WDeepLabv3+ is more complete than that segmented by DeepLabv3+.
These results illustrate that WDeepLabv3+ performs better on the detail regions and the similar objects.

\section{Conclusion}
Our proposed general DWT and IDWT layers are applicable to various wavelets,
which can be used to extract and process the data details during the network inference.
We design WaveSNets (WSegNet and WDeepLabv3+) by replacing the down-sampling with DWT
and replacing the up-sampling with IDWT, in U-Net, SegNet, and DeepLabv3+.
Experimental results on the CamVid, Pascal VOC, and Cityscapes show that
WaveSNets could well recover the image details and perform better for segmenting similar objects than their vanilla versions.

\bibliographystyle{named}
\bibliography{ijcai20}

\end{document}